\pdfoutput=1

\PassOptionsToPackage{dvipsnames}{xcolor}

\makeatletter
\newcommand*\betweenScriptAndFootnotesize{%
    \@setfontsize\betweenScriptAndFootnotesize{7.9pt}{9.5pt}%
}
\makeatother

\documentclass[11pt]{article}
\usepackage[table]{xcolor}

\usepackage[final]{acl}

\usepackage{times}
\usepackage{latexsym}
\usepackage{kotex}
\usepackage{graphicx}
\graphicspath{{./Images/}}
\usepackage{multirow}
\usepackage{comment}
\usepackage{hyperref}
\usepackage{amsmath}
\usepackage{amssymb}
\usepackage{bbm}
\usepackage{makecell}
\usepackage{listings}
\usepackage{tabularray}
\usepackage{colortbl}
\usepackage{soul}
\usepackage{subcaption}
\usepackage{pifont}
\usepackage[most]{tcolorbox}
\usepackage{enumitem}
\usepackage[colorinlistoftodos,prependcaption,textsize=scriptsize,textwidth=2cm]{todonotes}
\usepackage{mathrsfs}
\usepackage{dsfont}
\usepackage[normalem]{ulem}
    
\DeclareMathOperator*{\argmax}{arg\,max}
\DeclareMathOperator*{\argmin}{arg\,min}
\newcommand{\defEq}{\stackrel{.}{=}}
\newcommand{\pStar}{\mathbb{P}}

\definecolor{navyblue}{RGB}{213, 224, 244}
\definecolor{forestgreen}{RGB}{181,225,167}
\definecolor{dirt}{RGB}{253,219,110}
\definecolor{salmon}{RGB}{250, 219, 216}
\newtcbox{\highlight}[1]{%
    on line,
    boxsep=0pt,
    left=2pt,
    right=2pt,
    top=2pt,
    bottom=2pt,
    boxrule=0pt,
    arc=0pt,
    outer arc=0pt, 
    colback=#1!100 %
}

\usepackage[T1]{fontenc}

\usepackage[utf8]{inputenc}

\usepackage{microtype}

\usepackage{inconsolata}

\usepackage{adjustbox}
\usepackage{booktabs}
\usepackage{graphicx}
\usepackage{algorithm}
\usepackage{algorithmic}

\title{TRACT: Regression-Aware Fine-tuning Meets\\Chain-of-Thought Reasoning for LLM-as-a-Judge}

\author{Cheng-Han Chiang \\
  NTU  GICE \\
  Taiwan \\
  \texttt{dcml0714@gmail.com} \\\And
  Hung-yi Lee\\
  NTU GICE \\
  Taiwan\\
  \texttt{hungyilee@ntu.edu.tw} \\\And
  Michal Lukasik \\
  Google Research \\
  New York\\
  \texttt{mlukasik@google.com} \\}

\begin{document}
\maketitle

\begin{abstract}
The \emph{LLM-as-a-judge} paradigm uses large language models (LLMs) for automated text evaluation, which assigns a score to the text based on some scoring rubrics. 
Existing methods for LLM-as-a-judge use cross-entropy (CE) loss for fine-tuning, which neglects the numeric nature of score prediction. 
Recent work addresses numerical prediction limitations of LLM fine-tuning through regression-aware fine-tuning, which, however, does not consider chain-of-thought (CoT) reasoning for score prediction.
In this paper, we introduce \emph{TRACT} (\emph{Two-stage Regression-Aware fine-tuning with CoT}), a method combining CoT reasoning with regression-aware training. 
The training objective of TRACT combines the CE loss for learning the CoT reasoning and the regression-aware loss for the score prediction.
TRACT consists of two stages: first, a seed LLM is fine-tuned to generate CoTs; next, we retrain the seed LLM using the CoTs generated by the LLM trained in stage 1.
Experiments across four LLM-as-a-judge datasets and two LLMs show that TRACT significantly outperforms existing methods.
Extensive ablation studies validate the importance of each component in TRACT.\footnote{All code and models can be found at \url{https://github.com/d223302/TRACT}.}
\end{abstract}

\section{Introduction}
Large Language Models (LLMs) have been applied to evaluate written based on the \emph{fine-grained evaluation rubrics} specified in the input by outputting a score indicating the quality~\citep{chiang-lee-2023-large}.
The fine-grained evaluation rubrics define the criteria for scoring answers, and each instance to be evaluated can have a set of customized fine-grained evaluation rubrics.
This \emph{LLM-as-a-judge} paradigm has become the standard way to evaluate LLMs~\citep{zheng2023judging,dubois2024lengthcontrolledalpacaevalsimpleway,abdin2024phi3technicalreporthighly,lambert2025tulu3pushingfrontiers}, and has applications across diverse scenarios~\citep{10.5555/3692070.3694459,chiang-etal-2024-large}.
To induce the fine-grained assessment ability of LLMs, LLM-as-a-judge is typically trained with cross-entropy (CE) loss to predict chain-of-thought (CoT) reasoning about the evaluation, followed by a score~\citep{kim2024prometheus,kim-etal-2024-prometheus,li2024generative}.

\begin{figure}[t!]
    \centering
    \begin{subfigure}[b]{1.0\linewidth}
        \centering
        \includegraphics[clip, trim=20px 10px 122px 4px, width=1.0\linewidth]{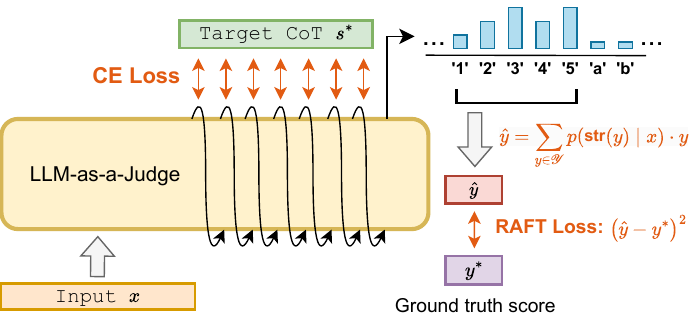}
        \caption{CoT-RAFT fine-tuning objective (\S\ref{ss:CoT-RAFT}).}
        \label{fig: CoT-RAFT}
    \end{subfigure}
    \hfill
    \begin{subfigure}[b]{1.0\linewidth}
        \centering
        \includegraphics[clip, trim=25px 10px 25px 8px, width=1.0\linewidth]{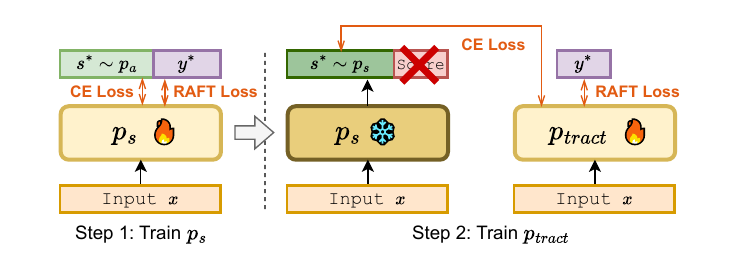}
        \caption{TRACT algorithm (\S\ref{subsection: Self-generated CoTs}).}
        \label{fig: TRACT algorithm}
    \end{subfigure}
    \caption{TRACT method overview. 
    (a) Illustration of the \emph{CoT-RAFT fine-tuning objective} (Eq.~\ref{eq: cot-raft full loss}), used in both stages of TRACT.
    (b) Two fine-tuning stages of TRACT (also see Algorithm~\ref{a:TRACT_ALGO}). 
    \emph{Stage 1}: model $p_{\rm s}$ is trained over the ground truth scores and the annotation CoTs (generated by the annotation model $p_{\rm a}$). 
    \emph{Stage 2}: CoT supervision is sampled from $p_{\rm s}$ (frozen at this stage) and used to fine-tune the final model $p_{\rm tract}$.}
    \label{fig:TRACT.pdf}
\end{figure}

Although fine-tuning LLM-as-a-judge with CE loss is a well-established practice~\citep{kim2024prometheus,kim-etal-2024-prometheus}, it has intuitive limitations when applied to predicting numerical targets.
For example, given a ground truth score '\emph{1}', placing high probability on the token '\emph{5}' is penalized the same as placing high probability on the token '\emph{2}', despite the fact that they induce very different errors in terms of numerical metrics such as squared error.
To mitigate this, \citet{raft} proposed regression-aware fine-tuning (RAFT; see \S\ref{ss:RAIL}), a method using squared error loss in fine-tuning LLMs.

RAFT has shown promising results on regression tasks; however, it does not incorporate CoT reasoning~\cite{wei2022chain,10.5555/3600270.3601883}. 
\citet{chiang-lee-2023-closer} show that not using CoT can harm the performance of LLM-as-a-judge.
This makes us ask: \textit{Can CoT reasoning be effectively integrated into regression-aware fine-tuning?}

In this paper, we propose \textbf{T}wo-stage \textbf{R}egression-\textbf{A}ware fine-tuning with \textbf{C}o\textbf{T} reasoning (\textbf{TRACT}), a method that combines the benefits of CoT reasoning with RAFT for improved numerical prediction abilities..
In \emph{stage 1}, TRACT fine-tunes the seed LLM over both the CoT supervision and the ground truth scores in the original training data. 
TRACT's fine-tuning objective (which we refer to as CoT-RAFT) is the sum of CE loss over the CoT, and the RAFT loss over score prediction (see Figure~\ref{fig: CoT-RAFT}).
In \emph{stage 2}, TRACT fine-tunes the seed LLM from the CoTs generated by the model trained in stage 1 using the same CoT-RAFT objective.
This two-stage pipeline is for reducing the distribution between the seed LLM and the CoT supervision used for fine-tuning in stage 2 (see Figure~\ref{fig: TRACT algorithm}).
We summarize TRACT in Algorithm~\ref{a:TRACT_ALGO} and contrast its formulation against baselines in Table~\ref{tbl:approaches_summary}.

We conduct experiments on five datasets and two LLMs and show that TRACT consistently outperforms baselines by a large margin, including Prometheus-2-7B~\citep{kim-etal-2024-prometheus}, which is the best model of equal size from prior works.
Our contributions are as follows:
\begin{enumerate}[label=(\roman*),leftmargin=20pt,itemsep=0pt,topsep=0pt]
\item We propose TRACT, a fine-tuning method that combines CoT reasoning with regression-aware loss to improve the numerical scoring ability of LLMs (\S\ref{s:method}).
\item We show that TRACT yields significant performance improvements over existing baselines for LLM-as-a-judge. 
We validate the importance of each component in TRACT with extensive ablations (\S\ref{section: Results}, \S\ref{section: Further Analysis}).
\item We show that TRACT, while trained on point-wise LLM-as-a-judge datasets, can work well on RewardBench~\citep{rewardbench}, a pairwise comparison dataset (\S\ref{section: Comparing Point-Wise LLM-as-a-Judge and Reward Models}).
\item We release our models. Our best model surpasses the prior state-of-the-art model of comparable size by 0.05 in Pearson's correlation when the inference-time compute is limited.
\end{enumerate}

\begin{table*}
    \centering
    \renewcommand{\arraystretch}{1.25}
\resizebox{\linewidth}{!}{
    \begin{tabular}{@{}lllll@{}}
        \toprule
         \textbf{Category} & \textbf{Approach}   & \textbf{Predictor function ${\hat{y}}(x)$}  & \textbf{Fine-tuning loss} & \textbf{Sec}\\
\toprule
\multirow{4}{*}{\shortstack[l]{Baselines}} 
 & Standard fine-tuning and decoding w/o CoT  & $\argmax_{y \in \mathcal{Y}} {{p}_{\rm}}(y \,|\, x)$ & $- \log {{p}_{\rm}}(y^* \,|\, x)$& \S\ref{ss:standard_nocot}\\
& RAIL zero-shot~\citep{lukasik-etal-2024-regression}   & $\sum_{y' \in \mathcal{Y}} y' \cdot {p}_{\rm}(y' | x)$ & N/A& \S\ref{ss:RAIL}\\
& \multirow{1}{*}{\shortstack[l]{RAFT~\cite{raft}}}   & $\sum_{y' \in \mathcal{Y}} y' \cdot {p}_{\rm}(y' | x)$ & $\left(\hat{y}(x) - y^* \right)^2$& \S\ref{ss:RAIL}\\
 & Standard fine-tuning and decoding w/ CoT  & $\argmax_{y \in \mathcal{Y}} p( \texttt{str}( y ) \,|\,[x, \hat{s}])$; $\hat{s} \sim p(\cdot \,|\, x)$ & $- \log {p}( [s^*, y^*] \,|\, x)$ & \S\ref{ss:standard_cot}\\
\midrule
\multirow{1}{*}{\shortstack[l]{Ours}} & TRACT   & $ \sum_{y \in \mathcal{Y}} p( \texttt{str}( y ) \,|\,[x, \hat{s}]) \cdot y$; $\hat{s} \sim p(\cdot \,|\, x)$ 
& Algorithm~\ref{a:TRACT_ALGO}& \S\ref{s:method}\\
\bottomrule
\end{tabular}
}
\caption{Approaches to autoregressive regression with decoder-based LLMs. 
Here,
$p( \cdot \,|\,x )$ denotes a distribution over possible outputs given an input string $x$,
and $\hat{y}( x ) \in \mathbb{R}$ a predictor function.
$\mathcal{Y}$ denotes the targets space (unless otherwise stated, $\{1, 2, 3, 4, 5\}$).}

\label{tbl:approaches_summary}
\end{table*}

\section{Background}
We first introduce the notation and next review previous works on regression-aware numerical output prediction with autoregressive models.

\subsection{Notation}
Let $\mathcal{V}$ denote a finite vocabulary of \emph{tokens}, 
$\mathcal{X} \subseteq \mathcal{V}^*$ be a set of \emph{inputs}  comprising strings of tokens, $\mathcal{S} \subseteq \mathcal{V}^*$ the set of CoT reasonings,
and $\mathcal{Y} \subset \mathbb{R}$ be a set of numerical \emph{targets}.
We assume that each $y \in \mathcal{Y}$ has a unique string representation ${\tt str}( y ) \in \mathcal{V}^*$.
Let $\pStar$ denote a ground-truth distribution over $\mathcal{X} \times \mathcal{S} \times \mathcal{Y}$, 
with the decomposition $\pStar( x, s, y ) = \pStar( x ) \cdot \pStar( s \,|\,x )\cdot \pStar( y \,|\,[x, s] )$, where $[x, s]$ denotes a string concatenation of $x$ and $s$.
Denote the training data by $D_{\rm train}\in(\mathcal{X}\times \mathcal{Y})^N$, and each element in ${D}_{\rm train}$ is composed of $(x, y^{*})$.
Denote the training data augmented with CoT by $D_{\rm train}^{\rm CoT}\in(\mathcal{X}\times \mathcal{S}\times \mathcal{Y})^N$, and each element in ${D}_{\rm train}^{\rm CoT}$ is composed of $(x, s^*, y^{*})$, where $s^*$ can come from the human or LLM annotation.

In this paper, we focus on LLM-as-a-judge application; in this case, the input ${x}\in\mathcal{X}$ consists of the instructions for LLM as a judge, the scoring rubrics and a sample to be evaluated.
The CoT reasoning $s$ contains an analysis of the sample in the inputs, and the target space is $\mathcal{Y} = \{1, 2, 3, 4, 5\}$.
An example input-output pair is shown in Table~\ref{tab:evaluation_prompt} in the Appendix.

We denote an LLM by the probability distribution that it models given the input, that is, $p(\cdot \,|\, x)$.
We use subscripts to denote specific LLM variants: $p_0$ represents the seed LLM before fine-tuning, and $p_{\rm a}$ represents the LLM annotator used to generate the CoTs in $D_{\rm train}^{\rm CoT}$.

In the following sections, we describe different methods for training and inference of predictor $\hat{y}$.

\subsection{Standard Fine-tuning and Decoding with No CoT}
\label{ss:standard_nocot}
Fine-tuning without CoT seeks to adapt a seed LLM $p_0(\cdot | x)$ to the target distribution $\mathbb{P}$ by minimizing
a suitable \emph{loss function} $\ell \colon \mathcal{Y} \times \Delta_{\mathcal{V}^*} \to \mathbb{R}$,
where $\Delta_{\mathcal{V}^*}$ denotes the set of distributions over a set ${V}^*$.
A standard choice of $\ell$ is the cross-entropy loss: $\ell_{\rm CE}( y^*, p ) = - \log {p}( \texttt{str}( y^* ) \,|\, x)$, where $\texttt{str}(\cdot)$ converts a numerical target $y^*$ to its string representation. 
Inference is typically conducted by \emph{standard (autoregressive) decoding}, approximately seeking the mode of the distribution $p( \cdot \,|\,x )$, i.e., $\hat{y}_{\text{mode}}(x) := \argmax_{y \in \mathcal{Y}}\, p(y \,|\,x)$.

\subsection{Regression-Aware Inference and Fine-tuning} \label{ss:RAIL} 

Recently, \citet{lukasik-etal-2024-regression} noted that standard autoregressive decoding in LLMs implicitly minimizes the 0-1 error metric, $m(y, \hat{y}) = \mathds{1}(y \neq \hat{y})$, making it suboptimal for regression metrics. 
Motivated by the minimum Bayes risk decoding framework~\citep{kumar-byrne-2004-minimum}, \citet{lukasik-etal-2024-regression} propose Regression-Aware Inference for Language models (RAIL). 
RAIL aims to determine the Bayes-optimal prediction by minimizing the expected loss: 
$\hat{y}_{\rm RAIL}(x) \defEq \argmin_{v \in \mathbb{R}} \mathbb{E}_{y \sim p(\cdot \,|\,x)}\left[ m(\texttt{num}(y), v) \right],$
where $m$ is the metric of interests and $\texttt{num}(\cdot)$ converts a string output to its numerical equivalent. 
For squared error loss, $m(y, \hat{y}) = (y - \hat{y})^2$, the optimal decision rule simplifies to the expected value of the numerical predictions. 
In the case of finite numerical targets $\mathcal{Y}$, this can be exactly calculated by scoring and averaging a set of candidate targets: 
\begin{align} \hat{y}_{\rm RAIL}(x) \defEq \sum_{y \in \mathcal{Y}} p(\texttt{str}(y) \,|\,x) \cdot y. \label{eq:RAIL_scoring} \end{align}

While RAIL is based on the Bayes-optimal decision rule, directly applying this rule to a model fine-tuned with CE loss can still result in high regression error. 
This is due to the misalignment between the CE fine-tuning loss and the regression evaluation metrics~\citep{raft}. 
To address this, \citet{raft} proposed incorporating the RAIL optimal decision rule directly into the fine-tuning process using squared loss, leading to the Regression-Aware Fine-Tuning (RAFT) loss: 
\begin{align} \ell_{\rm RAFT}(y^*, p) = \left(y^* - \hat{y}_{\rm RAIL}(x)\right)^2. \label{eq:bayes-opt-train2} \end{align}

\section{Method: TRACT}
\label{s:method}

While achieving notable success on natural language regression tasks such as STS-B~\citep{cer-etal-2017-semeval}, RAIL and RAFT do not use CoT reasoning. 
In this section, we start by discussing learning with CoT, and next introduce our method that incorporates CoT into RAIL and RAFT.

\subsection{Learning and Inference with CoT}
\label{ss:standard_cot}

LLM-as-a-judge typically decodes a CoT sequence prior to outputting the final score~\citep{chiang-lee-2023-closer}.
LLM-as-a-judge is fine-tuned over a dataset $D^{\rm CoT}_{\rm train} \in ( \mathcal{X} \times \mathcal{S} \times \mathcal{Y} )^N$ of $N$ (input, CoT, target score) tuples drawn from $\mathbb{P}$.
The CoT annotations $s^*$ often come from a powerful LLM (e.g. GPT-4); we denote the distribution of CoT annotations provided in the dataset as $p_{\rm a}(\cdot\,|x)$.
The empirical loss for fine-tuning is $\hat{L}_{\rm CoT}( p ) = \frac{1}{N} \sum_{( x, y^*, s^* ) \in D^{\rm CoT}_{\rm train}} \ell( y^*, s^*, p )$.

A typical choice is again the CE loss, i.e., %
    $\ell_{\rm CoT}( y^*, s^*, p ) = - \log {p}(  [s^*, y^*] | x)$.
Inference is typically done by generatively predicting a CoT reasoning and the score,
which samples from the distribution $p$, thus approximating the mode of the distribution: %
$\hat{y}_{\rm CoT~mode}( x ) \defEq \argmax_{y \in \mathcal{Y}} p( \texttt{str}( y ) \,|\,[x, \hat{s}])$; $\hat{s} \sim {p}( \cdot \,|\, x)$.

\subsection{Regression-Aware Fine-Tuning with CoT}
\label{ss:CoT-RAFT}
CoT reasoning elicits a reasoning $s$ before outputting the score $y$.
We propose and define the \emph{CR (CoT-RAIL)} predictor as first sampling a CoT $\hat{s}$ conditioning on the input $x$, and then applying the RAIL predictor when conditioning on $x$ and $\hat{s}$,
\begin{equation}
    \begin{aligned}
       \hat{y}_{\rm CR}(x) \,\defEq 
        \sum_{y \in \mathcal{Y}} p( \texttt{str}( y ) \,|\,[x, \hat{s}]) \cdot y; \hat{s} \sim p(\cdot \,|\,x)
    \end{aligned}
    \label{equation: CoT-RAIL with K}
\end{equation}
A practical question is how to determine where the CoT reasoning ends and where the score starts so that the RAIL predictor can be used at the appropriate position.
In practice, we assume each CoT ends with the string $d=$'\texttt{\highlight{forestgreen}{So the overall score is}}'.
We include the string $d$ at the end of each CoT in the training set to ensure that the fine-tuned model always ends its CoT with $d$.

In principle, we can sample a CoT and obtain the score prediction using Equation~\ref{equation: CoT-RAIL with K}, repeat the CoT sampling multiple times to estimate the score, and average the scores.
Unless stated otherwise, we sample $\hat{s}$ once, as shown in Equation~\ref{equation: CoT-RAIL with K}. %

\begin{algorithm*}[ht!]
   \caption{TRACT: Two-stage Regression-Aware fine-tuning with CoT reasoning}
      \label{a:TRACT_ALGO}
    \begin{algorithmic}[1]
    \STATE \textbf{input:} CoT annotations distribution $p_{\rm a}$, base model $p_0$, training data $\{(x, y^*)\}$, mixing coefficient $\lambda$.\\
    \STATE \textbf{stage 1:} train $p_{\rm s}$ using the objective $\ell_{\rm CoT-RAFT}^{\lambda}( y^*, p_{\rm a}, p )$ initializing from $p_0$\\
    \STATE \textbf{stage 2:} train $p_{\rm tract}$ using the objective $\ell_{\rm CoT-RAFT}^{\lambda}( y^*, p_{\rm s}, p )$ initializing from $p_0$\\
    \STATE \textbf{return} $p_{\rm tract}$
\end{algorithmic}
\end{algorithm*}

For training, analogously to the original RAFT algorithm, we use the CoT-RAIL predictor in the squared error loss. 
To guide the model in generating CoTs, we apply CE loss to the CoTs. %
We combine the two losses using a weighting coefficient $\lambda$, forming the \emph{CoT-RAFT} objective:
\begin{equation}
    \begin{aligned}
       \lefteqn{\ell_{\rm CoT-RAFT}^{\lambda}( y^*, p_{\rm t}, p ) = }\\
        & \lambda \left( \sum_{y \in \mathcal{Y}} p( \texttt{str}( y ) \,|\,[x, \hat{s}]) \cdot y - y^* \right)^2\\
        & - \log {p}( [\hat{s}, y^*] \,|\, x);  \hat{s} \sim p_{\rm t}(\cdot  \,|\, x),
    \label{eq: cot-raft full loss}
    \end{aligned}
\end{equation}
where $p_{\rm t}$ denotes the (target) model used to generate the CoTs for training, which can be $p_{\rm a}$ (GPT-4), but we explore other options next.

\subsection{Self-Generated Chain-of-Thoughts (CoTs)}
\label{subsection: Self-generated CoTs} 

When training with CoT-RAFT (Eq.~\ref{eq: cot-raft full loss}), the score predictor is conditioned on the CoT generated by a target model, denoted as $\hat{s} \sim p_{\rm t}(\cdot \,|\, x)$.
However, at inference time, CoT-RAIL (Eq.~\ref{equation: CoT-RAIL with K}) relies on self-generated CoTs from the fine-tuned model $p$, i.e., $\hat{s} \sim p(\cdot \,|\, x)$.
In our case, the CoTs used for training are from GPT-4 ($p_{\rm a}$) and may be very different from the CoTs generated by the fine-tuned model $p$.
This creates a mismatch between the CoTs used during training and inference, which can harm the performance, as empirically shown in \S\ref{section: Results}.

To mitigate the distribution shift in annotation and model CoTs, we propose to fine-tune the model using CoTs more closely aligned with those sampled from the trained model $p(\cdot \,|\, x)$. 
Let $p_{\rm s}$ denote the model initialized from the seed LLM $p_0$ and fine-tuned on the annotation CoTs sampled from $p_{\rm a}$.
For each input $x$ in the training data, we prompt $p_{\rm s}$ to generate a response containing a CoT and a score prediction. 
We then \emph{discard} the score prediction and pair the generated CoT with the corresponding original ground truth score, creating a new training dataset, denoted as $D_{\rm self}$. 
Formally, $D_{\rm self} = \{(x, \text{CoT}_{\rm s}(x), y^*) | (x, y^*) \in D_{\rm train}, \text{CoT}_{\rm s}(x) \sim p_{\rm s}(\cdot | x)\}$, where ${D}_{\rm train}$ is the original training dataset, $y^*$ is the ground truth score, and $\text{CoT}_{\rm s}(x)$ is the CoT generated by $p_{\rm s}$ for input $x$. 

We use $D_{\rm self}$ to train from the seed LLM $p_0$ to obtain the final TRACT model, $p_{\rm tract}$. 
The complete algorithm for training the TRACT model is shown in Algorithm~\ref{a:TRACT_ALGO} and Figure~\ref{fig:TRACT.pdf}.
In summary, we first train the seed LLM on the training data with the CoT annotation from $p_{\rm a}$ to obtain $p_{\rm s}$, use $p_{\rm s}$ to generate CoTs to form $D_{\rm self}$, and use $D_{\rm self}$ to retrain $p_{0}$ again, leading to the resulting model $p_{\rm tract}$.

\section{Experiment Setup}
\label{section: Experiment Setup}

\textbf{Models}\quad
We fine-tune TRACT from two LLMs, Mistral-7B-Instruct-v0.2~\citep{jiang2023mistral7b} and Llama-3.1-8B-Instruct~\citep{grattafiori2024llama3herdmodels}.

\noindent \textbf{Training Dataset}\quad
We follow \citet{kim-etal-2024-prometheus} and use \emph{Feedback Collection}~\citep{kim2024prometheus} as the training set.
The training set contains roughly 100K samples.
The whole dataset, including the samples to be evaluated, the evaluation responses, and ground truth scores, are generated by GPT-4.
We follow the procedure in \S\ref{subsection: Self-generated CoTs} to construct $D_{\rm self}$.
We analyze the quality of the self-generated CoTs in Appendix~\ref{appendix: Quality of Self-Generated CoTs}.

\noindent \textbf{Test Datasets}\quad
We use four datasets for point-wise LLM-as-a-judge, following \citet{kim-etal-2024-prometheus}.

\textbf{(1)} \emph{Feedback Bench}~\citep{kim2024prometheus}: This is the official test set of Feedback Collection.
Feedback Bench contains 1K responses to evaluate.
The instructions and scoring rubrics in Feedback Bench do not overlap with the training set.

\textbf{(2)} \emph{FLASK}~\citep{ye2024flask}: This is a fine-grained evaluation benchmark with 200 test prompts and 2000 responses from Alpaca-7B~\citep{alpaca}, Vicuna-13B, Bard~\citep{bard}, and GPT-3.5-Turbo-0613.

\textbf{(3)} \emph{Vicuna Bench}~\citep{vicuna2023}: This is a single-turn dialogue dataset with 80 user instructions.
\citet{kim2024prometheus} extends this dataset for LLM-as-a-judge by crafting the scoring rubrics for each user instruction and generating responses by WizardLM-13B~\citep{xu2024wizardlm}, Vicuna-13B~\citep{vicuna2023}, Llama-2-Chat-13B~\citep{touvron2023llama2openfoundation}, and ChatGPT-3.5-Turbo-0613.
The dataset contains 320 responses to evaluate.

\textbf{(4)} \emph{MT Bench}~\citep{zheng2023judging}: MT Bench is a multi-turn dialogue dataset with diverse tasks.
To form a direct assessment dataset, \citet{kim2024prometheus} prepare human-written rubrics and generate the dialogue responses from WizardLM-13B, Vicuna-13B, Llama-2-Chat-13B, and ChatGPT-3.5-Turbo-0613.
The dataset has 320 responses to evaluate.

\noindent \textbf{Baselines}\quad
We compare TRACT against multiple baselines that differ in their training objectives and in whether they use CoT reasoning, including:
(1) \emph{Standard fine-tuning and decoding without CoT} (\S\ref{ss:standard_nocot}), (2) \emph{Standard fine-tuning and decoding with CoT} (\S\ref{ss:standard_cot}), (3) \emph{Zero-shot RAIL}: Using RAIL predictor (Equation~\ref{eq:RAIL_scoring}) on the seed LLM $p_0$, (4) \emph{RAFT} (\S\ref{ss:RAIL}).
A detailed comparison of the baselines is shown in Table~\ref{tbl:approaches_summary}.

We also compare with (5) \emph{Prometheus-2-} \emph{7B}~\citep{kim-etal-2024-prometheus}, a model obtained by merging the model trained on Feedback-Collection and another model trained on Preference Collection~\citep{kim-etal-2024-prometheus}, a pairwise ranking dataset.
\citet{kim-etal-2024-prometheus} show that merging the models trained on two datasets significantly improves the performance.
Prometheus-2-7B is trained from Mistral-7B-Instruct-v0.2, the same seed LLM used in this paper, and it is the best open-source model of the same size for LLM-as-a-judge currently.

\begin{table*}[ht!]
    \centering
    \betweenScriptAndFootnotesize{}
    \begin{adjustbox}{max width=\textwidth}
    \begin{tabular}{ccccc|cccccccc|cc}
    \toprule
          \multicolumn{5}{c|}{\textbf{Train/Inference Configuration}}  &  \multicolumn{2}{c}{\textbf{\underline{FB Bench}}}& \multicolumn{2}{c}{\textbf{\underline{FLASK}}}& \multicolumn{2}{c}{\textbf{\underline{Vic. Bench}}}&\multicolumn{2}{c|}{\textbf{\underline{MT Bench}}}&  \multicolumn{2}{c}{\textbf{\underline{Average}}}\\
          Id&CoT&  Train& Data & Inf.&   $r$& $\rho$& $r$& $\rho$& $r$&$\rho$&  $r$&  $\rho$&  $r$ &  $\rho$\\
          \midrule
          \multicolumn{15}{c}{\textit{\textbf{Baselines}}} \\

B.1 &\textbf{\ding{55}}&  CE& GPT-4$^\ddagger$ & Decode&  0.890&   0.891& 0.355& 0.361& 0.429& 0.414& 0.279&  0.268&  0.488&  0.483\\
B.2 &\textbf{\ding{51}}&  CE& GPT-4&  Decode&  0.872 & 0.872 & 0.413 & 0.407 & 0.463 & 0.456 & 0.480 & 0.482 & 0.557 & 0.554 \\

B.3 &\textbf{\ding{55}}&  \textbf{\ding{55}}& \textbf{\ding{55}} &  RAIL&  0.197 & 0.175 & 0.200 & 0.149 & 0.281 & 0.165 & 0.309 & 0.216 & 0.247 & 0.176 \\

B.4 &\textbf{\ding{55}}&  RAFT& GPT-4$^\ddagger$ &  RAIL&  \textbf{0.932} & \textbf{0.930} & \underline{0.509} & \textbf{0.502} & \underline{0.567} & \underline{0.519} & 0.483 & 0.469 & \underline{0.623} & \underline{0.605} \\

\midrule
\multicolumn{15}{c}{\textit{\textbf{Not directly comparable models (Different Training Data)}}} \\
B.5 &\textbf{\ding{51}}&  \multicolumn{2}{c}{Prometheus-2-7B$^\dagger$} &  Decode&  0.845 & 0.847 & 0.512 & 0.493 & 0.488 & 0.480 & \underline{0.519} & \underline{0.483} & 0.591 & 0.576 \\
6 &\textbf{\ding{51}}&  \multicolumn{2}{c}{CLoud (Reward model)} &  Decode$^\ast$&  0.381& 0.376& 0.228& 0.168& 0.229& 0.311
& 0.511& 0.506& 0.337& 0.340\\
\midrule
\multicolumn{15}{c}{\textit{\textbf{TRACT (ours)}}} \\
7 & \textbf{\ding{51}}& C-RAFT & Self& C-RAIL & \underline{0.931} & \underline{0.930} & \textbf{0.518} & \underline{0.501} & \textbf{0.593} & \textbf{0.552} & \textbf{0.555} & \textbf{0.529} & \textbf{0.650} & \textbf{0.628} \\
\midrule
\multicolumn{15}{c}{\textit{\textbf{Ablation analysis for TRACT}}} \\
A.1 & \textbf{\ding{51}}& C-RAFT& GPT-4& C-RAIL & 0.879 & 0.880 & 0.418 & 0.419 & 0.528 & 0.513 & 0.399 & 0.418 & 0.556 & 0.558 \\
A.2 & \textbf{\ding{51}}& CE & Self& C-RAIL & 0.919 & 0.917 & 0.468 & 0.436 & 0.562 & 0.526 & 0.517 & 0.503 & 0.617 & 0.596 \\
A.3 & \textbf{\ding{51}}& CE & Self& Decode& 0.873 & 0.873 & 0.358 & 0.346 & 0.418 & 0.404 & 0.435 & 0.426 & 0.521 & 0.512 \\
A.4 & \multicolumn{4}{c|}{TRACT with stage 2 initialized from $p_{\rm s}$} & 0.674 & 0.684 & 0.448 & 0.437 & 0.505 & 0.477 & 0.432 & 0.421 & 0.515 & 0.505 \\
       
          \bottomrule
    \end{tabular}
    \end{adjustbox}
    \caption{The results for Mistral-7B-Instruct.
    The best and second-best results (excluding ablations) for each column are marked with \textbf{boldface} and \underline{underline}, respectively. 
    Explanation of abbreviations: \textit{Train}: training objective; \textit{Data}: source of CoT used for training; \textit{Inf.}: inference method; \textit{FB Bench}: Feedback Bench; \textit{Vic. Bench}: Vicuna Bench; $r$: Pearson's correlation coefficient; $\rho$: Spearman's rank correlation coefficient; \textit{C-RAFT}: CoT-RAFT; \textit{C-RAIL}: CoT-RAIL.
    $\dagger$: Prometheus-2-7B is obtained by merging two models trained from Feedback Collection and Preference Collection.
    We re-run the inference using the model released by~\citet{kim-etal-2024-prometheus} with the official code.
    $\ddagger$: When training without CoTs, the training target is simply the score, which is still generated by GPT-4.
    $\ast$: CLoud (trained from Llama-3-8B) is a reward model; it is not designed for LLM-as-a-judge benchmarks (Section~\ref{section: Comparing Point-Wise LLM-as-a-Judge and Reward Models}).} %
    \label{tab:mistral main result}
\end{table*}

\noindent \textbf{Training and Evaluation}\quad
We fine-tune the LLMs with LoRA~\citep{hu2022lora} using Llama-Factory~\citep{zheng2024llamafactory}.
We set $\lambda=1$ in Equation~\ref{eq: cot-raft full loss} unless otherwise stated.

During inference, for each sample in a dataset, we predict a score from the LLM using either standard decoding or the predictor in Equation~\ref{eq:RAIL_scoring} or \ref{equation: CoT-RAIL with K}, and calculate the correlation coefficient against the ground truth scores in the datasets.
We report Pearson's $r$~\citep{pearson1895vii}, Spearman's $\rho$~\citep{spearman1961proof}, and Kendall's $\tau$~\citep{kendall1938new} (see Kendall's $\tau$ results in Appendix~\ref{appendix: Results of Kendal's}; we find the trends to be consistent with other metrics).

\section{Main Results}
\label{section: Results}

The results for Mistral are presented in Table~\ref{tab:mistral main result}, and the results for Llama are in Table~\ref{tab: llama3 main result} in the Appendix.
The first five rows (B.1$\sim$B.5) in the two tables correspond to the five baselines in \S\ref{section: Experiment Setup} in the same order.
We next discuss the main observations.

\noindent \textbf{TRACT outperforms standard training and decoding with CoT.}\quad
Standard training and decoding with CoT (row B.2) is the default training and inference for LLM-as-a-judge in most prior works~\citep{kim2024prometheus, kim-etal-2024-prometheus}.
For both Mistral and Llama, TRACT (row 7) outperforms this baseline by a significant margin.
When the base model is Mistral, TRACT reaches Pearson's $r$ as high as 0.650, significantly outperforming training and inference with CoT (row B.2), which only attains Pearson's $r$ of 0.557.
When using Llama, TRACT improves the average Pearson's $r$ by 0.100.

\noindent \textbf{TRACT outperforms prior equally-sized SoTA.}\quad
Across all datasets, TRACT, trained without additional data, consistently outperforms Prometheus-2-7B (row B.5), the prior SoTA of the same size, by an average of 0.059 Pearson's $r$.
While we use self-generated data, this differs from Prometheus-2-7B, which is further trained on Preference Collection, a dataset generated by GPT-4.

\noindent \textbf{TRACT outperforms RAFT.}\quad
RAFT (row B.4) can be considered as an ablation of TRACT that removes CoT.
TRACT outperforms RAFT, showing that CoT helps the model predict the score.
Quite surprisingly, RAFT achieves the best performance among all baselines.
RAFT even outperforms Prometheus-2-7B (row B.5) by 0.032 Pearson's $r$, which is trained over additional data corresponding to a ranking task.
The same observation holds for the models trained using Llama-3.1.

Next, we ablate each component of TRACT.

\noindent \textbf{Training on self-generated CoTs is critical.}\quad
Row A.1 corresponds to training with CoT-RAFT on the GPT-4 generated data, and running inference with CoT-RAIL, without further training on self-generated CoTs.
The average Pearson's $r$ without training on self-generated data is 0.094 lower than TRACT.
This shows that self-generated CoT is an important component of TRACT.

\noindent \textbf{Using CE for training harms the performance.}\quad
Row A.2 is the result when we replace the fine-tuning objective from CoT-RAFT with the standard CE loss but still use CoT-RAIL in inference.
This configuration leads to a Pearson's $r$ 0.033 lower than TRACT.
This shows that using CoT-RAFT for fine-tuning better aligns the fine-tuning objective and the predictor used during inference.

\noindent \textbf{CoT-RAFT objective is necessary: self-generated CoTs alone are insufficient.}\quad
Prior research has demonstrated that fine-tuning LLMs on self-generated data using the CE loss can be beneficial~\citep{yang-etal-2024-self}.
Here, we investigate whether CE loss fine-tuning on self-generated data alone can improve the performance compared with training on GPT-4 CoTs.
Contrary to expectations, standard CE loss fine-tuning on self-generated CoTs (row A.3) yielded a significantly lower average Pearson's $r$ of 0.512, far below TRACT's performance. 
Moreover, this approach performed worse than fine-tuning on GPT-4 generated data (row B.2). 
This shows that self-generated CoTs are harmful when fine-tuned with CE loss but are beneficial when combined with the CoT-RAFT objective.
Our finding marks the difference between our work and prior work using self-generated CoTs with CE loss, highlighting the distinct role of self-generated CoTs in TRACT.

\noindent \textbf{Stage 2 training needs to be initialized from the seed LLM.}\quad
In Algorithm~\ref{a:TRACT_ALGO}, the model in stage 2 is initialized from the seed LLM $p_{0}$.
As an alternative, we explore initializing from $p_{\rm s}$, the model trained in stage 1, while keeping everything else the same as TRACT.
The result shown in row A.4 achieves only 0.515 Pearson's $r$.
This shows that initialization in stage 2 from the seed LLM (as Algorithm~\ref{a:TRACT_ALGO} indicates) is crucial.

\section{Analysis}
\label{section: Further Analysis}
We conduct further analysis of the key components of TRACT to justify its design.

\subsection{On the Distribution Shift between Annotation vs Self-Generated CoTs}
\label{subsection: What Makes Self-Generated CoT Special}

In \S\ref{subsection: Self-generated CoTs}, we motivated using self-generated CoTs by the distribution shift between training and inference CoTs. 
To empirically demonstrate this shift, we design the following experiment.

Let ${D}_{\rm train}^{\rm CoT}(p_{\rm t}) = \{(x_i, s_i^{*} \sim p_{\rm t}(\cdot | x_i), y_i^*)\}_{i=1}^N$ represent the training dataset, where $x_i \in \mathcal{X}$ is the input, $s_i^{*} \in \mathcal{S}$ is the CoT sampled from model $p_{\rm t}$ used as the training data, and $y_i^* \in \mathcal{Y}$ is the ground truth score. 
Let $p(\cdot)$ denote the probability distribution defined by the LLM trained on ${D}_{\rm train}^{\rm CoT}(p_{\rm t})$. 
After $p$ is trained, we generate CoTs $\hat{s}_i \sim p(\cdot | x_i)$ for each input $x_i$ in ${D}_{\rm train}$, and predict scores conditioned on these CoTs, denoted as $\hat{y}_i(\hat{s}_i) \sim p(\cdot | x_i, \hat{s}_i)$. 
Using the fine-tuned model $p$, we can also predict a score conditioned on $s^{*}_i$, the CoT in the training data ${D}_{\rm train}^{\rm CoT}(p_{\rm t})$; this score is denoted as $\hat{y}_i({s}^*_i) \sim p(\cdot | x_i, {s}^*_i)$. 

\begin{table}[t!]
    \centering
    \resizebox{\linewidth}{!}{%
    \begin{tabular}{c|c|c|cc}
    \toprule
  \multirow{2}{*}{TRACT} & $s^*$: CoTs used  &  $\hat{s}$: CoTs sampled from  & \multicolumn{2}{c}{RMSE w/ $y^*$} \\
                         & for training      &  fine-tuned model  &  $\hat{y}(s^*)$& $\hat{y}~(\hat{s})$  \\
    \midrule
         Stage 1  & $s^*\sim p_{\rm a}$ &  $\hat{s}\sim p_{\rm s}$ & 0.12 & 0.63 \\
          Stage 2 & $s^*\sim p_{\rm s}$& $\hat{s}\sim p_{\rm tract}$ & 0.45 & 0.45 \\
    \bottomrule
    \end{tabular}
    }
    \caption{RMSE between predicted scores $\hat{y}$ and ground truth scores $y^*$ over the training data.
    Each row corresponds to stage 1 or stage 2 of TRACT.
    The CoT used for fine-tuning the model in each stage is denoted by $s^*$ and is generated by the model indicated in the second column.
    After fine-tuning, we sample $\hat{s}$ from the fine-tuned model; the fine-tuned model is specified in the third column
    The last two columns are the RMSE across two pairs of predictions: 1) the predicted scores $\hat{y}_i(s_i^*)$ and the true scores $y_i^*$,  and 2) the predicted scores $\hat{y}_i(\hat{s}_i)$ and the true scores $y_i^*$.
    Within each row, both  $\hat{y}_i(s_i^*)$ and $\hat{y}_i(\hat{s}_i)$ are predicted by the same fine-tuned model (indicated in the third column).
    }
    \label{tab:memorization and generalization}
\end{table}

To quantify how different $s_i^*$ and $\hat{s}_i$ are, we compare the root mean square error (RMSE) across two pairs of scores: 1) the predicted scores $\hat{y}_i(s_i^*)$ and the ground truth scores $y_i^*$, and 2) the predicted scores $\hat{y}_i(\hat{s}_i)$ and the ground truth scores $y_i^*$.
Notably, both $\hat{y}_i(\hat{s}_i)$ and $\hat{y}_i(s_i^*)$ are generated by the same LLM $p(\cdot)$ conditioned on the same input $x_i$ in the training set; the only distinction is the CoT: $\hat{y}_i(s_i^*)$ conditioned on $s_i^*$, and $\hat{y}_i(\hat{s}_i)$ conditioned on $\hat{s}_i$. 

The results are presented in Table~\ref{tab:memorization and generalization}. 
The first row shows the model $p_{\rm s}$ trained in stage 1 in TRACT, which is trained using CoTs sampled from $p_{\rm a}$ (GPT-4). 
We observe a significant RMSE gap of 0.51 between $\hat{y}_i(s_i^*)$ and $\hat{y}_i(\hat{s}_i)$.
This shows that for $p_{\rm s}$, the distribution of training CoTs and self-generated CoTs is very different.

The second row shows the result of the model fine-tuned in stage 2, $p_{\rm tract}$, which is trained on CoTs generated by $p_{\rm s}$.
When predicting scores using $p_{\rm tract}$, the RMSE gap between using the training CoTs $s^*$ and predicted CoTs $\hat{s}$ is 0. 
This indicates that for $p_{\rm s}$ (row 1), predicting scores using self-generated CoTs is more challenging than using CoTs generated by GPT-4, showing a substantial distribution shift between self-generated and GPT-generated CoTs. 
Conversely, for $p_{\rm tract}$, the prediction difficulty is similar for both types of CoTs, suggesting that the training and self-generated CoT distributions are closer.

The above results can also be interpreted as $p_{\rm s}$ memorizing the training CoTs and cannot generalize to its own generated CoTs, while $p_{\rm tract}$ is capable of generalizing to its own generations.

\begin{figure}[t!]
    \centering
    \includegraphics[clip, trim = 0px 0px 0px 0px,width=1.0\linewidth]{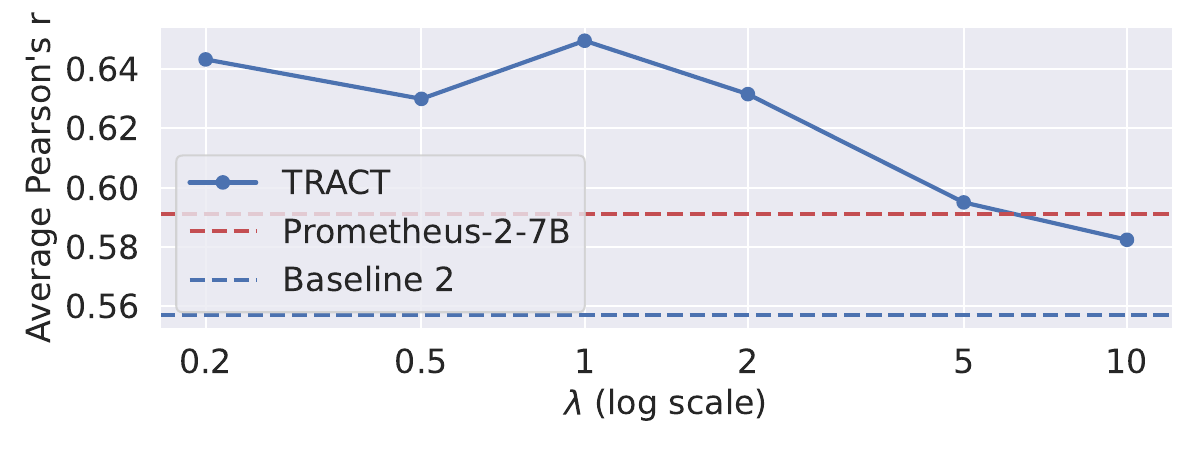}
    \caption{Performance of TRACT across varying values of $\lambda$ in Equation~\ref{eq: cot-raft full loss}. Results from the Mistral model. For a wide range of $\lambda$ values we find TRACT outperforming the baselines.}
    \label{fig:lambda.pdf}
\end{figure}

\subsection{Sensitivity to $\lambda$}
\label{subsection: Sensitivity to lambda}
We analyze the performance of TRACT when varying $\lambda$ in Equation~\ref{eq: cot-raft full loss} over a grid of values $\{0.2, 0.5, 1.0,2, 5, 10\}$, using Mistral as the seed model.
The results are shown in Figure~\ref{fig:lambda.pdf}.
We observe that $\lambda=1$ performs well. %
For most values of $\lambda$ TRACT outperforms both 1) the model trained using CE loss on the same dataset, and 2) Prometheus, the SoTA model of the same size.

\subsection{Multi-Objective Fine-tuning vs. Sequential Single-Objective Fine-tuning}
\label{subsection: Multitask Learning VS Continual Learning}
TRACT relies on the CoT-RAFT objective from Eq.~\ref{eq: cot-raft full loss}, combining the CE loss over CoTs and the RAFT loss over score predictions.
An alternative training strategy could involve a sequential, two-stage fine-tuning process: first, fine-tuning with the CE loss over both CoTs and scores (corresponding to baseline 2); and next, fine-tuning on the same dataset using solely the RAFT objective over scores.
In the second stage, while the model's output still includes CoTs, the CE loss is not computed for them, i.e., the second term in Eq.~\ref{eq: cot-raft full loss} is not used.

We empirically find that this sequential fine-tuning approach significantly degrades the model's ability to generate coherent and complete CoTs.
The resulting model struggles to terminate CoT outputs correctly and sometimes yields very long CoTs.
We show some example CoTs in Table~\ref{tab:corrupted CoTs} in Appendix~\ref{section: Corrupted CoTs}.
This deficiency highlights the critical role of multi-objective fine-tuning: jointly optimizing both the CE loss for CoTs and the squared-error loss for scores.

\begin{figure}[t!]
    \centering
    \includegraphics[clip, trim = 0px 0px 0px 0px,width=1.0\linewidth]{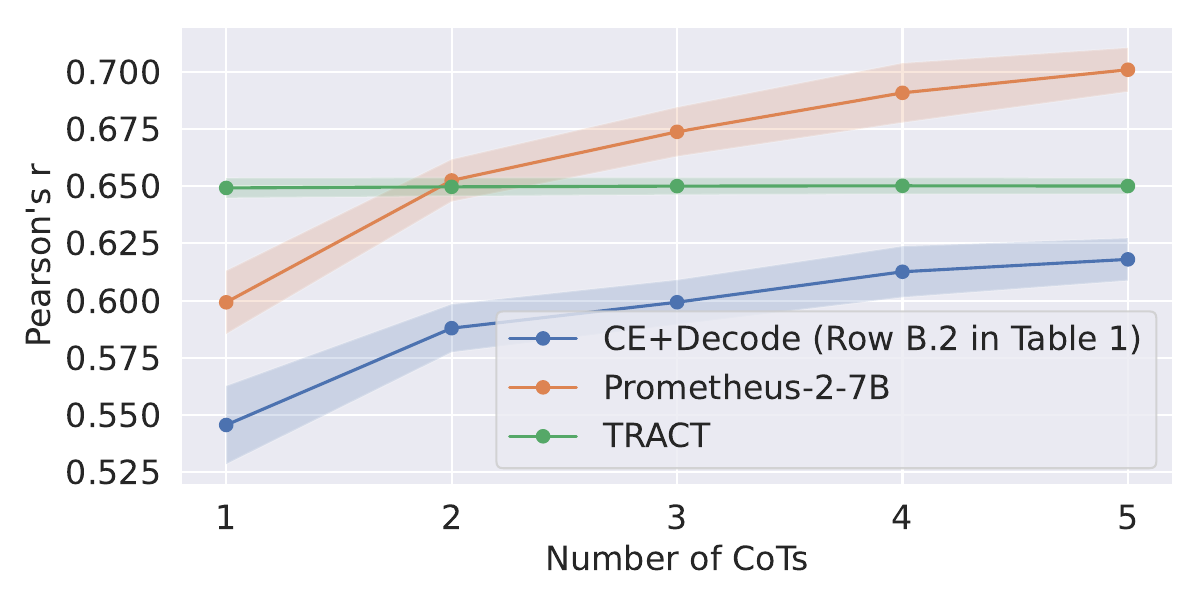}
    \caption{Average Pearson's $r$ as a function of the number of sampled CoTs. 
    Results from fine-tuning the Mistral model.
    Shaded regions correspond to the standard deviations across multiple inference runs with varying random seeds.
    Note that Prometheus is trained on significantly more data compared to other two methods in this Figure. Despite that, under limited inference budget, TRACT outperforms Prometheus.}
    \label{fig:n_cot.pdf}
\end{figure}

\subsection{Scaling the Number of Sampled CoTs}
\label{subsection: Scaling the Number of Sampled CoTs}
We analyze the effect of sampling multiple CoTs for TRACT during inference; this is done by sampling multiple $\hat{s}$, calculating $\hat{y}_{\rm CR}$ for each CoT (Eq.~\ref{equation: CoT-RAIL with K}), and averaging the $\hat{y}_{\rm CR}$.
For comparison, we also sample multiple CoTs and average the scores for the following two baselines: the model trained using the CE loss (baseline 2, row B.2 in Table~\ref{tab:mistral main result}), and Prometheus-2-7B.
To gauge the significance of the differences across methods, we repeat the experiment by varying the random seed.
For each seed, we sample $K$ CoTs to calculate the score, average the score across four testing dataset, and repeat the above process with four seeds.

The results are shown in Figure~\ref{fig:n_cot.pdf}.
We find that the performance of TRACT is stable across different numbers of CoTs, while the performance of standard decoding increases when sampling more CoTs.
While this might indicate TRACT cannot benefit from scaling the inference-time compute~\citep{irvine2023rewarding,brown2025large,snell2025scaling}, the superiority of TRACT under limited inference-time compute is clear.
Baseline 2 uses the same amount of training data and serves as a fair comparison baseline to TRACT.
Compared to it, TRACT with one CoT consistently outperforms baseline 2 with more than one CoTs.
Additionally, the variance of the performance of TRACT across the random seeds is much smaller than that for standard decoding.

TRACT outperforms Prometheus when sampling fewer CoTs.
Given that Prometheus is trained with more data, we find it to be a very strong outcome that TRACT outperforms Prometheus under the limited CoT sampling setting.

\section{Comparing Point-Wise LLM-as-a-Judge and Reward Models}
\label{section: Comparing Point-Wise LLM-as-a-Judge and Reward Models}

Given an instruction and a response, a reward model (RM) assigns a scalar reward to indicate the quality of the response~\citep{ArmoRM,liu2024skywork,dorka2024quantile}.
While this formulation is similar to pointwise LLM-as-a-judge, RMs are typically trained to evaluate specific attributes, e.g., helpfulness, harmlessness, etc.~\citep{bai2022training}.
As a result, RMs cannot take fine-grained evaluation metrics and evaluate based on the evaluation rubrics.
Furthermore, the reward of a RM is only meaningful when comparing the reward of two responses for the same instruction, since RMs are trained with pairwise data. 
In contrast, the score of point-wise LLM-as-a-judge is meaningful standalone since it directly corresponds to the description of a specific score in the scoring rubrics.

Critique-out-loud (\emph{CLoud})~\citep{ankner2024critique} RM is an RM that generates some CoT reasoning before predicting the reward using a regression head.
While CLoud uses self-generated CoT reasoning to improve the numeric prediction (the reward), there are several differences between CLoud and TRACT.
First, CLoud is trained on pairwise reward modeling datasets, while TRACT is trained on a point-wise LLM-as-a-judge dataset.
Next, TRACT uses the language model head to predict the score, while CLoud uses a regression head to predict the score.
Most importantly, CLoud, as a reward model, cannot take fine-grained evaluation rubrics and evaluate the responses based on the evaluation criteria, which is different from all other LLM-as-a-Judge based baselines in Section~\ref{section: Experiment Setup}, which can take fine-grained evaluation rubrics.

\subsection{Results on Point-wise LLM-as-a-Judge Datasets}
\label{subsection: Results on Point-wise LLM-as-a-Judge Datasets}
Since CLoud also uses CoT to improve numeric predictions, we compare CLoud with TRACT on point-wise LLM-as-a-judge datasets to understand whether RMs can perform well on point-wise LLM-as-a-judge datasets.
In row 7 in Table~\ref{tab:mistral main result}, we show the results of CLoud on LLM-as-a-judge datasets.
We can see that CLoud performs poorly, and TRACT significantly outperforms CLoud.
This is not surprising, as CLoud is a reward model trained on pairwise ranking datasets, and it cannot take the fine-grained evaluation metrics.
Our goal here is not to criticize the performance of CLoud; we use the above evidence to highlight that RMs are not the same as point-wise LLM-as-a-judge models.

\subsection{Results on RewardBench}
Having seen that RM cannot yield reasonable performance on point-wise LLM-as-a-judge datasets, one may ask whether the reverse holds as well: 
\textit{Can point-wise LLM-as-a-judge models be used as a reward model?}
To answer this question, we use \emph{RewardBench}~\citep{rewardbench}, a reward model dataset, to evaluate our baseline models and TRACT.
Each evaluating instance in RewardBench consists of an instruction, a winning response, and a losing one; the goal is to see whether a reward model can assign a higher reward to the winning response.
To test point-wise LLM-as-a-judge models on RewardBench, we prompt the model with both responses separately, obtaining scores for them, and then compare the scores for the final ranking.
We report the accuracy on RewardBench.

The results for RewardBench are presented in Table~\ref{tab:rewardbench} in the Appendix.
We see that TRACT performs quite well: the model fine-tuned from Mistral reaches an average score of 0.736, slightly worse than CLoud (0.759).
This shows that even though TRACT, a point-wise LLM-as-a-judge model, is not trained on pairwise ranking for reward modeling, it can still obtain reasonably good performance on RewardBench, a pairwise ranking benchmark. 
This is in stark contrast with the result of CLoud on point-wise LLM-as-a-judge shown in Table~\ref{tab:mistral main result}, where CLoud underperforms TRACT by a large margin.
Lastly, we note that TRACT, a model trained on only a point-wise dataset, outperforms Prometheus-2-7B, a model trained on point-wise scoring and pairwise ranking datasets.

\section{Related Work}
\subsection{LLM-as-a-Judge}
\label{subsection: LLM-as-a-Judge}
LLM-as-a-judge paradigm is an application of LLMs as an automatic judge to assess texts~\citep{chiang-lee-2023-large}.
There are two types of LLM-as-a-judge:
(1) \textbf{Direct assessment}: the LLM assigns a score to a given sample~\citep{chiang-lee-2023-large,liu-etal-2023-g}.
(2) \textbf{Pairwise ranking}: the LLM determines which sample is better given a pair of samples~\citep{zheng2023judging,wang2024pandalm,li2024generative}.
In this paper, we focus on improving the direct assessment ability of the LLM.

Many prior works focus on fine-tuning open-source LLMs into better LLM judges using open-source data~\citep{vu-etal-2024-foundational,li2024generative}.
LLM-as-a-judge is often trained with the CE loss over CoT reasoning and a score~\citep{kim2024prometheus,kim-etal-2024-prometheus}.

\subsection{Training LLMs with Self-Generated Data}
\label{subsection: self-generated}
An important component of TRACT is fine-tuning over self-generated CoTs.
Using self-generated data for training LLMs has been considered in the literature~\citep{wang-etal-2023-self-instruct,kim-etal-2023-aligning} and is sometimes referred to as \textit{self-training}~\citep{singh2024beyond} or \textit{self-distillation}~\citep{yang-etal-2024-self}.
Early work has already shown that self-distillation can improve the performance of machine translation~\citep{freitag2017ensembledistillationneuralmachine,guo2021selfdistillationmixuptrainingnonautoregressive}.
Self-generated data can be used to augment the training data of LLMs~\citep{wu2025metarewarding,sun2024easytohard} to mitigate the scarcity of high-quality labeled data.
Self-generated CoTs, when combined with proper filtering methods, have been shown to improve LLM's reasoning ability~\citep{zelikman2022star,gulcehre2023reinforcedselftrainingrestlanguage,dou-etal-2024-rest}.

Most related to our work is \citet{yang-etal-2024-self}, who argue that self-generated data helps fine-tuning by reducing the distribution gap between the training data and the seed LLM.
\citet{ren-etal-2024-learn} show that self-generated data yields lower perplexity under the seed LLM, which can explain the performance gain of fine-tuning on self-generated data.
While we confirm that self-generated data has a lower perplexity under the seed LLM (Appendix~\ref{subsection: Training data perplexity}), we use a fine-tuning objective different from prior works.
Interestingly, we find that fine-tuning over self-generated data with the CE loss, as recommended in prior works, does not improve the performance.
Our proposed objective does significantly benefit from self-generated CoTs, showing how self-generated data is important for TRACT.

\section{Conclusion}
\label{section: conclusion}
In this paper, we introduce Two-stage Regression-Aware fine-tuning with CoT reasoning (TRACT), a method integrating CoT reasoning with the RAFT framework to harness the step-by-step reasoning ability of LLMs while learning the numerical structure of the scoring task.
TRACT consists of (1) CoT-RAFT for training, (2) CoT-RAIL for inference, and (3) two-stage fine-tuning for self-generated CoTs training.
Our experiments across two models and five datasets show that TRACT consistently outperforms all prior baselines finetuned on the same dataset by a significant margin.
Under limited inference compute, TRACT also outperforms Prometheus-2-7B, a same-sized SoTA model trained on more data.
We carry out careful ablations to explicate the importance of each component in TRACT.
We believe our proposed method and open-source models can benefit the community working on LLM-as-a-judge.

\section*{Limitations}
One of the limitations of TRACT is that it requires access to the model's output probabilities, which is not always possible for proprietary models.
Our technique can be readily applied to open-source models.

\section*{Acknowledgments}
We thank the reviewers for their valuable and constructive feedback.
We compare with CLoud and include the results of RewardBench based on the reviewers' suggestions.
Cheng-Han Chiang is supported by a Google PhD Fellowship and a Ph.D. scholarship program by Delta Electronics.

\bibliography{custom}

\appendix

\section{Hyperparameters}
\label{appendix:Hyperparameters}
\paragraph{Training Hyperparameters}
We train our models using Llama-Factory~\citep{zheng2024llamafactory}.
All models are trained with LoRA~\citep{hu2022lora}.
We use all the default LoRA parameters in Llama-Factory: the rank of LoRA is set to 8, and we apply LoRA on all the linear layers of the transformer model.
We fine-tune the model with a learning rate of $1.0\times 10^{-5}$, use a cosine learning rate scheduler~\citep{loshchilov2017sgdr}, and fine-tune for two epochs.
The warm-up ratio is set to $1.0$.
We use bf16 for training.
The effective batch size of training is set to 8.
All our training is run on a single NVIDIA RTX A6000; training the model on the whole training dataset takes about 50 hours.
When training TRACT, we need to first fine-tune on the GPT-4 generated data and fine-tune on the self-generated data, making the total training time about 100 hours.

\paragraph{Inference Hyperparameters}
For inference, we use vLLM~\citep{kwon2023efficient}.
Most inference hyperparameters follow \citet{kim-etal-2024-prometheus}: the top-$p$ is set to $0.9$~\citep{Holtzman2020The}, the temperature is set to $1.0$, the repetition penalty is set to $1.03$, the maximum number of output tokens is set to $1024$.

The only different hyperparameter from \citet{kim-etal-2024-prometheus} is we sample only 1 CoT in Table~\ref{tab:mistral main result}, while they sample 3 CoTs and average the score.
To obtain the results of Prometheus-2-7B with 1 CoT, we use their official model from Huggingface~\citep{wolf-etal-2020-transformers} and use the official code from \url{https://github.com/prometheus-eval/prometheus-eval/tree/main/eval}.
We made some necessary modifications to the official code to make it executable; we find that we can reproduce the Prometheus results using our modified code.

\section{Supplementary Results}

\subsection{Quality of the Self-Generated CoTs}
\label{appendix: Quality of Self-Generated CoTs}
In this section, we investigate the quality of the self-generated CoTs used in the second training stage in TRACT.
To investigate the quality of self-generated CoTs, we use LLM-as-a-meta-judge to evaluate the evaluation CoTs generated by GPT-4 (the original CoTs in Feedback Collection~\citep{kim2024prometheus}) and the student model $p_{\rm s}$.
We call GPT-4 the \textbf{meta-judge} since it is used to evaluate the evaluation of LLM-as-a-judge.
We prompt the meta-judge to evaluate the quality of the evaluation CoTs using a 5-point scale, with 1 being the lowest and 5 the highest.
The evaluation prompt is shown in Table~\ref{tab:self-generated CoT evaluation}.
We ask meta-judge to evaluate whether the evaluation CoT follows the scoring rubric and is grounded on the response.

Note that the \texttt{[judgement]} to be evaluated in Table~\ref{tab:self-generated CoT evaluation} includes the score, which is the original score in Feedback Collection.
That is, no matter whether the evaluation CoT is generated from GPT-4 or $p_{\rm s}$, the final score is always the one in Feedback Collection. 
This is the setting that is used to train the models.
Including the score in this meta-evaluation experiment allows the meta-judge to determine if the final score is consistent with the evaluation CoTs.

We randomly sample 200 samples from the training set and collect the evaluation CoTs in the Feedback Collection (GPT-4 generated) and the CoTs generated by $p_{\rm s}$, the student model fine-tuned on GPT-4 CoTs.
The average score for GPT-4 generated CoTs is \textbf{4.78}/5.00, while the average score of CoTs generated by $p_{\rm s}$ is \textbf{4.50}/5.00.
While the average score of $p_{\rm s}$-generated is slightly worse than that of GPT-4, a score of 4.50/5.00 is still very high, indicating that the self-generated CoTs have good quality.

\begin{table*}[t]
    \tiny
    \centering

    \begin{subtable}[t]{0.95\textwidth}
        \centering
        \begin{tabular}{p{0.95\textwidth}}
        \toprule
        
\texttt{An instruction (might include an Input inside it), a response to evaluate, a reference answer that gets a score of 5, and a score rubric representing a evaluation criteria are given.}\\
\texttt{1. Write a detailed feedback that assess the quality of the response strictly based on the given score rubric, not evaluating in general.}\\
\texttt{2. After writing a feedback, write a score that is an integer between 1 and 5. You should refer to the score rubric.}\\
\texttt{3. The output format should look as follows: "Feedback: (write a feedback for criteria) [RESULT] (an integer number between 1 and 5)"}\\
\texttt{4. Please do not generate any other opening, closing, and explanations.}\\
\texttt{}\\
\texttt{\#\#\#The instruction to evaluate:}\\
\texttt{You are a well-known psychiatrist who has a reputation for being empathetic and understanding. A client comes to you saying they've had a really hard day at work. They describe their boss as being overly critical and not listening to their ideas, which has left them feeling frustrated and undervalued. They also mention feeling overwhelmed with the workload and a growing sense of loneliness as they are new to the city. How do you respond to them to make them feel heard and understood, as well as offering helpful advice?}\\
\texttt{}\\
\texttt{\#\#\#Response to evaluate:}\\
\texttt{It's indeed challenging to deal with a difficult boss and to carry a large workload, especially when you are new to a city and haven't established a supportive network. I would suggest taking a step back to think about your situation and perhaps speak to your boss regarding the issues you're facing. On the other hand, dealing with a large workload can be managed by prioritizing tasks, and maybe you can discuss it with your superiors too.}\\
\texttt{}\\
\texttt{In regards to feeling lonely, you might want to explore activities or groups that align with your interests to meet new people. And also, remember to take time for yourself to relax and destress. I'm here to assist you as we navigate through these issues.}\\
\texttt{}\\
\texttt{\#\#\#Reference Answer (Score 5):}\\
\texttt{I hear that you're dealing with quite a few challenging situations at once. It must be tough to have your ideas dismissed by your boss and to feel undervalued in your work environment. It's understandable that such experiences can lead to frustration and a feeling of being overwhelmed.}\\
\texttt{}\\
\texttt{It's important to acknowledge these feelings and not to dismiss them. Feeling heard and validated in our experiences can often be the first step towards resolving the challenges we face. So, I want you to know that your feelings are completely valid and it's okay to be frustrated and overwhelmed.}\\
\texttt{}\\
\texttt{In terms of dealing with your boss, have you considered seeking a private conversation with them to express your feelings? It's often beneficial to communicate what you're experiencing, using "I" statements to express how you feel when your ideas are not taken into consideration. This can lead to mutual understanding and possibly a change in behavior.}\\
\texttt{}\\
\texttt{About the workload, it might help to prioritize tasks and potentially delegate, if possible. Also, consider discussing your workload with your superiors. There might be a misunderstanding about what's manageable or they might have suggestions about how to handle the situation.}\\
\texttt{}\\
\texttt{On the personal front, feeling lonely, especially when you're new to a city, can be really hard. Seek out opportunities to meet new people, perhaps through hobbies, community activities, or online groups. It might take a bit of time, but gradually, you can build a network of friends and acquaintances.}\\
\texttt{}\\
\texttt{Remember, it's perfectly okay to have bad days and it's important to take care of your mental health. Consider incorporating activities into your daily routine that make you happy and help you unwind. This could be anything from reading, yoga, going for a walk, or even listening to your favorite music.}\\
\texttt{}\\
\texttt{Please know that you're not alone in this. I'm here to support you through this challenging time and together, we can work towards resolving these issues.}\\
\texttt{}\\
\texttt{\#\#\#Score Rubrics:}\\
\texttt{[Is the model able to identify and react correctly to the emotional context of the user's input?]}\\
\texttt{Score 1: The model utterly fails to grasp the user's emotional context and responds in an unfitting manner.}\\
\texttt{Score 2: The model sporadically identifies the emotional context but frequently replies in a manner that doesn't match the user's emotional status.}\\
\texttt{Score 3: The model typically identifies the emotional context and reacts suitably, but occasionally misreads or misjudges the user's feelings.}\\
\texttt{Score 4: The model often identifies the emotional context and reacts suitably, with minor cases of misreading or misjudging.}\\
\texttt{Score 5: The model flawlessly identifies the emotional context of the user's input and consistently responds in a considerate and empathetic manner.}\\
\texttt{}\\
\texttt{\#\#\#Feedback:}\\
        \bottomrule
        \end{tabular}
        \caption{Input example with the input prompt template.}
        \label{subtab:input_prompt}
    \end{subtable}

    \vspace{0.1cm} %

    \begin{subtable}[t]{0.95\linewidth}
        \centering
        \begin{tabular}{p{0.95\columnwidth}}
        \toprule
        \texttt{The response offers practical, original, and effective solutions (...). \highlight{forestgreen}{So the overall score is} 3}\\
        \bottomrule
        \end{tabular}
        \caption{Example output}
        \label{subtab:example_output}
    \end{subtable}

    \caption{The input prompt template (with an input example) and an example output used in our main experiment.}
    \label{tab:evaluation_prompt}
\end{table*}

\begin{table}[t]
    \betweenScriptAndFootnotesize{}
    \centering

    \begin{subtable}[t]{0.95\linewidth}
        \centering
        \begin{tabular}{p{0.95\linewidth}}
        \toprule
        \texttt{A judge model is required to evaluate the quality of a response generated by an AI model based on the user's instruction.}\\ \\
        \texttt{The following are the instructions and inputs for the judge model:} \\ \\
        \texttt{\#\#\#  Begin of Inputs to Judge Model }\\
        \texttt{[input]} \\
        \texttt{\#\#\# End of Inputs to Judge Model }\\ \\
        
        \texttt{Here is the judgement from the Judge model:} \\ \\ 
        \texttt{\#\#\#  Begin of Judgement }\\
        \texttt{[judgement]} \\
        \texttt{\#\#\# End of Judgement }\\ \\
        \texttt{Question: Please evaluate the quality of the judgement based on whether the judgement is grounded on the responses and carefully follows the rubric. Provide some reasoning and analysis to support your evaluation. Next, provide an **integer rating** between 1 to 5, where 1 is the lowest quality and 5 is the highest quality, to asssess the quality of the judgement. Please always conclude with "Score: score", where score is the integer rating.}\\
        \bottomrule
        \end{tabular}
    \end{subtable}

    \vspace{0.1cm} %

    \caption{The prompt used to evaluate the quality of the evaluation CoTs.
    The \texttt{[input]} includes all the input to the LLM-as-a-judge model that is used to generate the evaluation CoT, as shown in Table~\ref{tab:evaluation_prompt} (Appendix).
    The \texttt{[judgement]} includes the evaluation CoT and the final score.}
    \label{tab:self-generated CoT evaluation}
\end{table}
 
\subsection{Results of Kendal's $\tau$ Correlation Coefficients}
\label{appendix: Results of Kendal's}

The results of Kendall's $\tau$ correlation coefficient when fine-tuned with Mistral are shown in Table~\ref{tab:mistral Kendall tau}.

\begin{table*}[ht!]
\betweenScriptAndFootnotesize{}
    \centering
    \begin{tabular}{ccccc|cccc|c}
    \toprule
          \multicolumn{5}{c|}{\textbf{Train/Inference Configuration}}  &  \textbf{\underline{FB Bench}}& {\textbf{\underline{FLASK}}}& {\textbf{\underline{Vic. Bench}}}&\textbf{\underline{MT Bench}}&  {\textbf{\underline{Average}}}\\
          Id&CoT&  Train& Data & Infer.&   $\tau$ & $\tau$ & $\tau$ & $\tau$ \\
          \midrule
          \multicolumn{10}{c}{\textit{\textbf{Baselines}}} \\

          B.1 &\textbf{\ding{55}}&  CE& GPT-4$^\ddagger$ & Decode&   \textbf{0.824} & 0.294 & 0.344 & 0.211 & 0.418 \\
B.2 &\textbf{\ding{51}}&  CE& GPT-4&  Decode&  0.798 & 0.328 & 0.380 & \underline{0.380} & 0.472 \\
B.3 &\textbf{\ding{55}}&  \textbf{\ding{55}}& \textbf{\ding{55}} &  RAIL&  0.130 & 0.109 & 0.122 & 0.150 & 0.127 \\
B.4 &\textbf{\ding{55}}&  RAFT& GPT-4$^\ddagger$ &  RAIL&  0.818 & \underline{0.375} & 0.401 & 0.342 & 0.484 \\
\midrule
B.5 &\textbf{\ding{51}}&  \multicolumn{2}{c}{Prometheus-2-7B$^\dagger$} &  Decode&  0.765 & \textbf{0.405} & 0.411 & 0.392 & \underline{0.493} \\
\midrule
\multicolumn{10}{c}{\textit{\textbf{TRACT (ours)}}} \\
7 & \textbf{\ding{51}}& C-RAFT & Self& C-RAIL & \underline{0.820} & 0.373 & \textbf{0.423} & 0.386 & \textbf{0.500} \\
\midrule
\multicolumn{10}{c}{\textit{\textbf{Ablation analysis for TRACT}}} \\
A.1 & \textbf{\ding{51}}& C-RAFT& GPT-4& C-RAIL & 0.763 & 0.315 & 0.409 & 0.307 & 0.448 \\
A.2 & \textbf{\ding{51}}& CE & Self& C-RAIL & 0.798 & 0.321 & 0.406 & 0.369 & 0.474 \\
A.3 & \textbf{\ding{51}}& CE & Self& Decode&  0.795 & 0.281 & 0.339 & 0.337 & 0.438 \\

          \bottomrule
    \end{tabular}
    \caption{The Kendall's $\tau$ results for Mistral-7B-Instruct.
    The best and second-best results (excluding ablations) for each column are marked with \textbf{boldface} and \underline{underline}, respectively. 
    Explanation of abbreviations: \textit{Train}: training objective; \textit{Data}: source of CoT used for training; \textit{Inf.}: inference method; \textit{FB Bench}: Feedback Bench; \textit{Vic. Bench}: Vicuna Bench; \textit{C-RAFT}: CoT-RAFT; \textit{C-RAIL}: CoT-RAIL.
    $\dagger$: Prometheus-2-7B is obtained by merging two models trained from Feedback Collection and Preference Collection.
    We rerun the inference using the model released by~\citet{kim-etal-2024-prometheus} with the official code.
    $\ddagger$: When training without CoTs, the training target is simply the score, which is still generated by GPT-4.}
    \label{tab:mistral Kendall tau}
\end{table*}

\subsection{Results from Llama-3.1}
We report the results from experiments with Llama-3.1-8B-Instruct in Table~\ref{tab: llama3 main result} and the corresponding Kendall's $\tau$ in Table~\ref{tab:llama Kendall tau}.
We find that TRACT outperforms all baseline methods.
Compared with Prometheus-2-7B in Table~\ref{tab:mistral main result}, TRACT fine-tuned on Llama-3.1 8B leads to much stronger results.
However, Prometheus-2-7B is fine-tuned with a different base model (Mistral-7B), which is a smaller model, and thus the results may not be directlu comparable.

\subsection{Perpexity of Training Data}
\label{subsection: Training data perplexity}
We use the base LLM $p_{0}$ (Mistral-7b-Instruct-v0.2) to calculate the perplexity of the CoTs from $p_{\rm a}$ and the CoTs generated by $p_{\rm s}$ and $p_{\rm tract}$.
For the above three types of CoTs, their perplexity is 7.78, 6.67, and 6.34 respectively.
Similar to~\citet{ren-etal-2024-learn}, we find that self-generated CoTs (from $p_{\rm s}$ and $p_{\rm tract}$) yield lower perplexity compared to those generated by GPT-4.

\begin{table*}[ht!]
    \betweenScriptAndFootnotesize{}
    \centering
    \begin{adjustbox}{max width=\textwidth}
    \begin{tabular}{lcccc|cccccccc|cc}
        \toprule
        \multicolumn{5}{c|}{\textbf{Train/Inference Configuration}}  &  \multicolumn{2}{c}{\textbf{\underline{FB Bench}}} & \multicolumn{2}{c}{\textbf{\underline{FLASK}}} & \multicolumn{2}{c}{\textbf{\underline{Vic. Bench}}} & \multicolumn{2}{c|}{\textbf{\underline{MT Bench}}} &  \multicolumn{2}{c}{\textbf{\underline{Average}}} \\
        Id   & CoT      & Train   & Data    & Inf    & $\rho$ & $r$       & $\rho$ & $r$       & $\rho$ & $r$       & $\rho$ & $r$       & $\rho$ & $r$       \\
        \midrule
        \multicolumn{15}{c}{\textit{\textbf{Baselines}}} \\
        B.1  & \ding{55} & CE      & GPT-4   & Decode & 0.857   & 0.857     & 0.433   & 0.435     & 0.423   & 0.400     & 0.556   & 0.541     & 0.567   & 0.558     \\
        B.2  & \ding{51} & CE      & GPT-4   & Decode & 0.834   & 0.835     & 0.484   & 0.475     & 0.483   & 0.467     & 0.494   & 0.466     & 0.574   & 0.561     \\
        B.3  & \ding{55} & \ding{55} & \ding{55} & RAIL   & 0.689   & 0.683     & 0.445   & 0.412     & 0.487   & 0.485     & 0.583   & 0.547     & 0.551   & 0.532     \\
        B.4  & \ding{55} & RAFT    & GPT-4$^\ddagger$ & RAIL   & \textbf{0.918} & \underline{0.920} & \underline{0.493} & \textbf{0.506} & \underline{0.541} & \underline{0.509} & \underline{0.614} & \underline{0.618} & \underline{0.642} & \underline{0.639} \\
        \midrule
        \multicolumn{15}{c}{\textit{\textbf{TRACT (ours)}}} \\
        7    & \ding{51} & C-RAFT  & Self    & C-RAIL & \underline{0.917} & \textbf{0.920} & \textbf{0.493} & \underline{0.500} & \textbf{0.650} & \underline{0.605} & \textbf{0.639} & \textbf{0.672} & \textbf{0.675} & \textbf{0.674} \\
        \bottomrule
    \end{tabular}
    \end{adjustbox}
    \caption{The results when using Llama3.1-8B as the base model.
    The best and second-best results for each column are marked with \textbf{boldface} and \underline{underline}, respectively.
    The abbreviations are the same as those in Table~\ref{tab:mistral main result}.
    $\ddagger$: When training without CoTs, the training target is simply the score, which is still generated by GPT-4.}
    \label{tab: llama3 main result}
\end{table*}

\begin{table*}[ht!]
\betweenScriptAndFootnotesize{}
    \centering
    \begin{tabular}{ccccc|cccc|c}
    \toprule
          \multicolumn{5}{c|}{\textbf{Train/Inference Configuration}}  &  \textbf{\underline{FB Bench}}& {\textbf{\underline{FLASK}}}& {\textbf{\underline{Vic. Bench}}}&\textbf{\underline{MT Bench}}&  {\textbf{\underline{Average}}}\\
          Id&CoT&  Train& Data & Infer.&   $\tau$ & $\tau$ & $\tau$ & $\tau$ \\
          \midrule
          \multicolumn{10}{c}{\textit{\textbf{Baselines}}} \\

          B.1 &\textbf{\ding{55}}&  CE& GPT-4$^\ddagger$ & Decode&   0.772 & 0.353 & 0.333 & 0.429 & 0.472 \\
B.2 &\textbf{\ding{51}}&  CE& GPT-4&  Decode&  0.749 & \textbf{0.385} & \underline{0.406} & 0.372 & 0.478 \\
B.3 &\textbf{\ding{55}}&  \textbf{\ding{55}}& \textbf{\ding{55}} &  RAIL&  0.535 & 0.298 & 0.360 & 0.398 & 0.398 \\
B.4 &\textbf{\ding{55}}&  RAFT& GPT-4$^\ddagger$ &  RAIL&  \underline{0.799} & \underline{0.371} & 0.386 & \underline{0.455} & \underline{0.503} \\
\midrule
\multicolumn{10}{c}{\textit{\textbf{TRACT (ours)}}} \\
7 & \textbf{\ding{51}}& C-RAFT & Self& C-RAIL & \textbf{0.805} & 0.367 & \textbf{0.468} & \textbf{0.494} & \textbf{0.534}  \\
          
          \bottomrule
    \end{tabular}
    \caption{The Kendall's $\tau$ results for Mistral-7B-Instruct.
    The best and second-best results for each column are marked with \textbf{boldface} and \underline{underline}, respectively. 
    Explanation of abbreviations: \textit{Train}: training objective; \textit{Data}: source of CoT used for training; \textit{Inf.}: inference method; \textit{FB Bench}: Feedback Bench; \textit{Vic. Bench}: Vicuna Bench; \textit{C-RAFT}: CoT-RAFT; \textit{C-RAIL}: CoT-RAIL.
    $\ddagger$: When training without CoTs, the training target is simply the score, which is still generated by GPT-4.}
    \label{tab:llama Kendall tau}
\end{table*}

\section{Corrupted CoTs for Section~\ref{subsection: Multitask Learning VS Continual Learning}}
\label{section: Corrupted CoTs}
We show examples of corrupted CoTs generated from the model trained with sequential fine-tuning (see Section~\ref{subsection: Multitask Learning VS Continual Learning}).
In the first sample, the evaluation CoT is short, and it is unclear what the '[5]' means here.
In the second sample, the model generates a string '(Score 5)' in the middle, and fails to generate the required output in expected format.
In the third example, the fine-tuned model generates '(Score 2)', which also do not follow the expected output format.

\begin{table*}[ht!]
    \centering
    \footnotesize
    \begin{tabular}{p{15cm}}
        \toprule
        Example CoTs\\
        \midrule
        The response effectively adapts its language to the user's Hindi language preference and rural Indian background by providing a step-by-step guide in Hindi. [5] So the overall score is 2 \\
        \midrule
        Feedback: The response effectively addresses the issue of improving a chatbot's ability to validate and incorporate user input. It outlines several strategies including the use of advanced Natural Language Processing techniques, handling multi-part inputs, maintaining conversation context, encouraging user feedback, and iterative testing. These approaches closely align with the reference answer. However, some enhancements could be made to provide a more concise and clearer explanation, especially in the context of the score rubric. (Score 5) This response demonstrates a high level of understanding of the problem domain and proposes practical solutions. The use of examples, such as \"breaking down multi-part questions into smaller components\" and \"maintaining a context stack,\" adds clarity to the suggestions and improves overall comprehension. By referring back to the original problem statement and directly addressing the criteria outlined in the rubric, the response effectively justifies its recommendations with strong evidence. So the overall score is 4\\
        \midrule
        Feedback: The model does not demonstrate an understanding of the cultural idiom 'an old broom knows the corners' and fails to provide a thoughtful response. Instead, it dismisses the metaphor as confusing and insignificant in the corporate context. Its reaction is insensitive and lacks the ability to relate the idiom to practical business scenarios, significantly diminishing the value of the exchange. (Score 2) So the overall score is 1\\
        \bottomrule
    \end{tabular}
    \caption{Examples of corrupted CoT sampled from a model trained via sequential fine-tuning: first fine-tuning using CE loss, and second fine-tuning with RAFT.
    See Section~\ref{subsection: Multitask Learning VS Continual Learning} for more details.}
    \label{tab:corrupted CoTs}
\end{table*}

\section{License}
\label{appendix: License}
We release the model under the Apache 2.0 license.
Since our models are trained using Feedback Collection, a dataset generated from OpenAI GPT-4, the models are further subject to \href{https://openai.com/policies/row-terms-of-use/}{OpenAI's Terms of Use}.
Additionally, the models fine-tuned from Llama-3.1-8B-Instruct is also further subject to the \href{https://github.com/meta-llama/llama-models/blob/main/models/llama3_1/LICENSE}{Llama 3.1 Community License}.

\begin{table}[ht]
    \centering
    \begin{adjustbox}{max width=\columnwidth}
    \begin{tabular}{lccccc}
        \toprule
        \textbf{Id} & \textbf{Chat} & \textbf{Chat-Hard} & \textbf{Safety} & \textbf{Reasoning} & \textbf{Average} \\
        \midrule
        B.1   & 0.592 & 0.290 & 0.700 & 0.355 & 0.484 \\
        B.2   & 0.629 & 0.305 & 0.120 & 0.368 & 0.355 \\
        B.3   & 0.595 & 0.358 & 0.651 & 0.312 & 0.479 \\
        B.4   & 0.883 & \textbf{0.581} & \textbf{0.807} & 0.828 & \textbf{0.775} \\
        \midrule
        B.5.1 & 0.679 & 0.423 & 0.673 & 0.378 & 0.538 \\
        B.5.2 & 0.855 & 0.491 & 0.771 & 0.765 & 0.720 \\
        B.6   & \textbf{0.965} & 0.455 & 0.754 & \textbf{0.862} & \underline{0.759} \\
        \midrule
        \multicolumn{6}{c}{\textit{\textbf{TRACT (ours)}}} \\
        7.M   & \underline{0.927} & 0.542 & 0.759 & 0.716 & 0.736 \\
        7.L   & 0.922 & 0.434 & \underline{0.799} & \underline{0.837} & 0.748 \\
        \midrule
        A.1   & 0.777 & 0.513 & 0.569 & 0.482 & 0.585 \\
        A.2   & 0.886 & \underline{0.564} & 0.773 & 0.645 & 0.717 \\
        A.3   & 0.564 & 0.292 & 0.385 & 0.194 & 0.359 \\
        A.4   & 0.824 & 0.469 & 0.700 & 0.568 & 0.640 \\
        \bottomrule
    \end{tabular}
    \end{adjustbox}
    \caption{Accuracy on RewardBench.
    The indices are identical to those in Table~\ref{tab:mistral main result}.
    7.M and 7.L denote TRACT trained from Mistral and Llama-3-8B, respectively.
    All models are trained from Mistral, except for B.6 (CLoud) and 7.L, which are based on Llama-3-8B.
    For Prometheus-2-7B, we report two results since it can be used as a pointwise (B.5.1) and a pairwise (B.5.2) LLM-as-a-judge.}
    \label{tab:rewardbench}
\end{table}

\end{document}